\documentclass[runningheads]{llncs}

\usepackage{graphicx}
\begin{document}
\title{Task Adaptive Pretraining of Transformers for Hostility Detection}
%
%
\author{Tathagata Raha\and
Sayar Ghosh Roy\and
Ujwal Narayan\and
Zubair Abid\and
Vasudeva Varma}
\authorrunning{Raha et al., 2021}
%
\institute{Information Retrieval and Extraction Lab (iREL)\leavevmode \\ International Institute of Information Technology, Hyderabad\leavevmode \\ \leavevmode \  
\email{\leavevmode \\ \{tathagata.raha, sayar.ghosh, ujwal.narayan, zubair.abid\}@research.iiit.ac.in, vv@iiit.ac.in}\\
}
\maketitle              
\begin{abstract}
Identifying adverse and hostile content on the web and more particularly, on social media, has become a problem of paramount interest in recent years. With their ever increasing popularity, fine-tuning of pretrained Transformer-based encoder models with a classifier head are gradually becoming the new baseline for natural language classification tasks. In our work, we explore the gains attributed to Task Adaptive Pretraining (TAPT) prior to fine-tuning of Transformer-based architectures. We specifically study two problems, namely, (a) Coarse binary classification of Hindi Tweets into Hostile or Not, and (b) Fine-grained multi-label classification of Tweets into four categories: hate, fake, offensive, and defamation. Building up on an architecture which takes emojis and segmented hashtags into consideration for classification, we are able to experimentally showcase the performance upgrades due to TAPT. Our system (with team name `iREL IIIT’) ranked first in the `Hostile Post Detection in Hindi’ shared task with an F1 score of 97.16\% for coarse-grained detection and a weighted F1 score of 62.96\% for fine-grained multi-label classification on the provided blind test corpora.

\end{abstract}

\section{Introduction}
With the increase in the number of active users on the internet, the amount of content available on the World Wide Web and more specifically, that on social media has seen a sharp rise in recent years. A sizable portion of the available content contains instances of hostility thereby posing potential adverse effects upon its readers. Content which is hostile in the form of, say, a hateful comment, unwarranted usage of offensive language, attempt at defaming an individual or a post spreading some sort of misinformation circulates faster as compared to typical textual information. \cite{mathew-etal-2019-hate,Vosoughi1146} Identifying and pinpointing such instances of hostility is of utmost importance when it comes to ensuring the sanctity of the World Wide Web and the well-being of its users and as such, multiple endeavours have been made to design systems which can automatically identify toxic content on the web. \cite{pink_2,pinna_fake,cyberbully,pink_1,hasoc20}

In this work, we focus on the problem of identifying certain Hindi Tweets which are hostile in nature. We further analyze whether the Tweet can fit into one or more of the following buckets: hateful, offensive, defamation, and fake. The popularity of pretrained Transformer-based \cite{vaswani2017attention} models for tasks involving Natural Language Understanding is slowly making them the new baseline for text classification tasks. In such a scene, we experiment with the idea of Task Adaptive Pretraining. \cite{gururangan2020don} IndicBERT \cite{kakwani2020indicnlpsuite}, which is similar to BERT \cite{devlin2018bert} but trained on large corpora of Indian Language text is our primary pretrained Transformer of choice for dealing with Hindi text.

We adopt a model architecture similar to Ghosh Roy et al., 2020 \cite{ghosh_roy_hate_speech_2020}, which leverages information from emojis and hashtags within the Tweet in addition to the cleaned natural language text. We are able to successfully portray 1.35\% and 1.40\% increases for binary hostility detection and on average, 4.06\% and 1.05\% increases for fine-grained classifications into the four hostile classes on macro and weighted F1 metrics respectively with Task Adaptive Pretraining (TAPT) before fine-tuning our architectures for classification.

\begin{table}
\centering
\caption{Distribution of Supervised labels in Training set}\label{stats}
\begin{tabular}{|l|l|}
\hline
\textbf{Label} & \textbf{Frequency}\ \ \ \  \\
\hline
Non-Hostile \ \ \ \ \ \ \  & 3050 \\
Defamation & 564 \\
Fake & 1144 \\
Hate & 792 \\
Offensive & 742 \\
\hline
\end{tabular}
\end{table}

\begin{table}
\centering
\caption{Distribution of labels in the Test set}\label{teststats}
\begin{tabular}{|l|l|}
\hline
\textbf{Label} & \textbf{Frequency}\ \ \ \  \\
\hline
Non-Hostile \ \ \ \ \ \ \  & 873 \\
Defamation & 169 \\
Fake & 334 \\
Hate & 234 \\
Offensive & 219 \\
\hline
\end{tabular}
\end{table}

\section{Dataset}

The dataset for training and model development was provided by the organizers of the Constraint shared task\footnote{\url{constraint-shared-task-2021.github.io}} \cite{patwa2021overview} \cite{bhardwaj2020hostility}. The data was in the form of Tweets primarily composed in the Hindi language and contained annotations for five separate fields. Firstly, a coarse grained label for whether the post is hostile or not was available. If a Tweet was indeed hostile, it would not carry the `not-hostile’ tag. Hostile Tweets carried one or more tags indicating it’s class of hostility among the following four non-disjoint sets (definitions for each class were provided by the Shared Task organizers)\footnote{Examples of cleaned Tweet text for each category from the provided dataset have been presented for illustration purposes only.}:

\begin{enumerate}
\item \textbf{Fake News}: A claim or information that is verified to be untrue.
\\
\item \textbf{Hate Speech}: A post targeting a specific group of people based on their ethnicity, religious beliefs, geographical belonging, race, etc., with malicious intentions of spreading hate or encouraging violence.
\\
\item \textbf{Offensive}: A post containing profanity, impolite, rude, or vulgar language to insult a targeted individual or group.
\\
\item \textbf{Defamation}: A mis\-information regarding an individual or group.
\end{enumerate}

A collection of 5728 supervised training examples were provided which we split into training and validation sets in a 80-20 ratio, while a set of 1653 Tweets served as the blind test corpora. The mapping from a particular class to its number of training examples have been outlined in table \ref{stats}. The distribution of labels within the test set is shown in table \ref{teststats}. Note that the test labels were released after the conclusion of the shared task. Throughout, a post marked as `non-hostile’ cannot have any other label while the remaining posts can theoretically have $n$ labellings, $n$ $\in$ $\{1, 2, 3, 4\}$.

\section{Approach}

In this section, we describe our model in detail and present the foundations for our experiments. We acknowledge that the language style for social media text differs from that of formal as well as day to day spoken language. Thus, a model whose input is in the form of Tweets should be aware of and be able to leverage information encoded in the form of emojis and hashtags. We base our primary architecture on that of Ghosh Roy et al., 2020 \cite{ghosh_roy_hate_speech_2020} with a few modifications.

\subsection{Preprocessing and Feature Extraction}

Similar to Ghosh Roy et al., 2020, \cite{ghosh_roy_hate_speech_2020} the raw input text is tokenized on whitespaces plus symbols such as commas, colons and semicolons. All emojis and hashtags are extracted into two separate stores. The cleaned Tweet text which is the primary information source for our model is free from non-textual tokens including smileys, URLs, mentions, numbers, reserved words, hashtags and emojis. The tweet-preprocessor\footnote{\url{github.com/s/preprocessor}} python library was used for categorizing tokens into the above mentioned classes.

To generate the centralised representation of all emojis, we utilize emoji2vec \cite{DBLP:journals/corr/EisnerRABR16} to generate 300 dimension vectors for each emoji and consider the arithmetic mean of all such vectors. We use the ekphrasis\footnote{\url{github.com/cbaziotis/ekphrasis}} library for hashtag segmentation. The segmented hashtags are arranged in a sequential manner separated by whitespaces and this serves as the composite hashtag or `hashtag flow' feature. Thus, we leverage a set of three features, namely, (a) the cleaned textual information, (b) the collective hashtag flow information, and (c) the centralised emoji embedding.

\begin{figure}
\includegraphics[width=\textwidth]{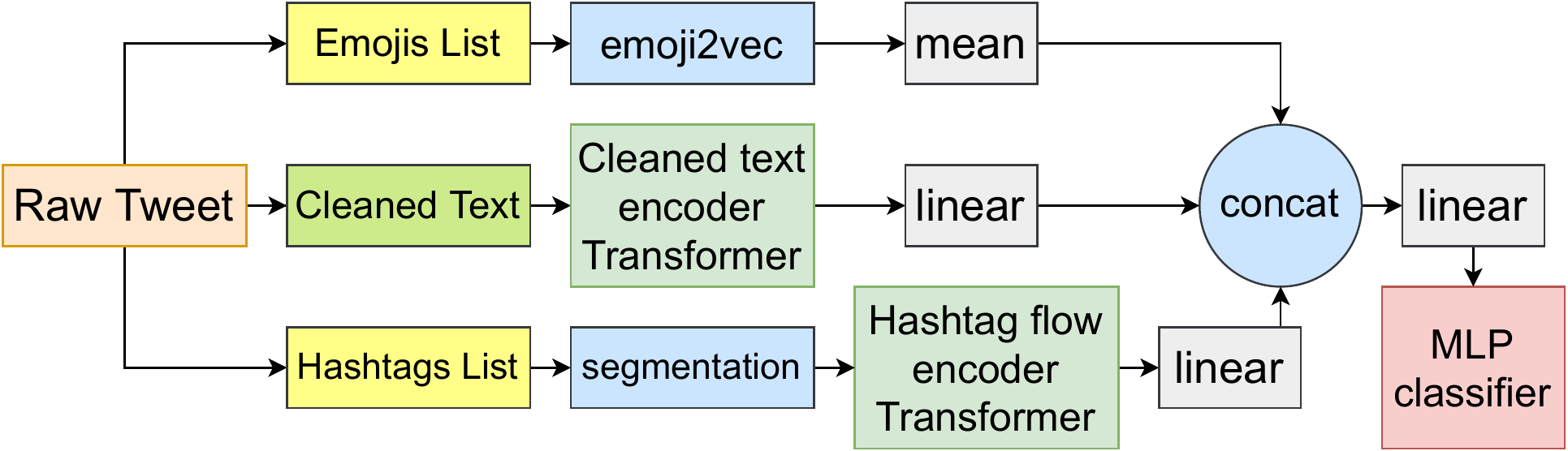}
\caption{Model Architecture} \label{arch}
\end{figure}

\subsection{Architecture}

In this subsection, we outline the overall flow of information pieces from the set of input features to label generation. We leverage two Transformer models to generate embeddings of dimension size 768 for each of the cleaned text and the hashtag flow features. Each of the two Transformer-based embeddings are passed through two linear layers to yield the final vector representations for cleaned text and hashtag collection. The three vectors: cleaned text, composite hashtag and centralised emoji representations are then concatenated and passed through a linear layer to form the final 1836-dimension vector used for classification. A dense multi-layer perceptron serves as the final binary classifier head. The overall information flow is presented in Figure \ref{arch}. For the multi-label classification task, we trained our architecture individually to yield four separate binary classification models. In all cases, we performed an end to end training on the available training data based on cross-entropy loss. 

\subsection{Task Adaptive Pretraining}

We turn to Gururangan et al., 2020, \cite{gururangan2020don} which showcases the boons of continued pretraining of Transformer models on natural language data specific to certain domains (Domain Adaptive Pretraining) and on the consolidated unlabelled task-specific data (Task Adaptive Pretraining). Their findings highlighted the benefits of Task Adaptive Pretraining (TAPT) of already pretrained Transformer models such as BERT on downstream tasks like text classification. We experimented with the same approach for our task of hostility detection in Hindi having IndicBERT as our base Transformer model. Our results (in section \ref{res}) clearly showcases the gains attributed to this further pre-training with the masked language modeling (MLM) objective. Note that only the cleaned text encoder Transformer model is the one undergoing TAPT. The hashtag sequence encoder Transformer is initialized to pretrained IndicBERT weights. We create a body of text using all of the available training samples and in that, we add each sample twice: firstly, we consider it as is i.e the raw Tweet is utilized, and secondly, we add the cleaned Tweet text. A pretrained IndicBERT Transformer is further pretrained upon this body of text with the MLM objective and we use these Transformer weights for our cleaned text encoder before fine-tuning our architecture on the training samples.

\begin{table}
\centering
\caption{Results on the Validation split for every category (\% Weighted F1 Scores)}\label{validation}
\begin{tabular}{|l|l|l|l|}
\hline
\textbf{Metric} & \textbf{Without TAPT}\ \ \ \  & \textbf{With TAPT}\ \ \ \ \ \ \  & \textbf{Gains}\ \ \ \ \\
\hline
Hostility (Coarse)\ \ \ \ \  & 96.87 & 98.27 & \ 1.40 \\
\hline
Defamation & 86.47 & 86.31 & -0.16\\
Fake & 89.53 & 90.99 & \ 1.46\\
Hate & 85.69 & 87.06 & \ 1.37\\
Offensive & 87.12 & 88.66 & \ 1.54\\
\hline
\end{tabular}
\end{table}

\begin{table}
\centering
\caption{Results on the Validation split for every category (\% Macro F1 Scores)}\label{validationmacro}
\begin{tabular}{|l|l|l|l|}
\hline
\textbf{Metric} & \textbf{Without TAPT}\ \ \ \  & \textbf{With TAPT}\ \ \ \ \ \ \  & \textbf{Gains}\ \ \ \ \\
\hline
Hostility (Coarse)\ \ \ \ \  & 96.84 & 98.19 & \ 1.35 \\
\hline
Defamation & 59.43 & 63.38 & \ 3.95\\
Fake & 83.69 & 86.52 & \ 2.83\\
Hate & 70.77 & 74.20 & \ 3.43\\
Offensive & 68.72 & 74.73 & \ 6.01\\
\hline
\end{tabular}
\end{table}

\begin{table}
\centering
\caption{Shared Task Results: Top 3 teams on public leaderboard (\% F1 Scores) }\label{taskres}
\begin{tabular}{|l|l|l|l|}
\hline
\textbf{Metric} & \textbf{iREL IIIT} (Us)\ \ \ \  & \textbf{Albatross}\ \ \ \ \ \ \ \ \ \   & \textbf{Quark}\ \ \ \ \ \ \ \ \ \ \ \   \\
\hline
Hostility (Coarse)\ \ \ \ \  & 97.16 & 97.10 & 96.91 \\
\hline
Defamation & 44.65 & 42.80 & 30.61 \\
Fake  & 77.18 & 81.40 & 79.15 \\
Hate & 59.78 & 49.69 & 42.82 \\
Offensive & 58.80 & 56.49 & 56.99 \\
\hline
Weighted (Fine) & 62.96 & 61.11 & 56.60 \\
\hline
\end{tabular}
\end{table}

\section{Results}
\label{res}
In tables \ref{validation} and \ref{validationmacro}, we present metrics computed on our validation set. We observe 1.35\% and 1.40\% increases in the macro and weighted F1 scores for binary hostility detection and on average, 4.06\% and 1.05\% increases in macro and weighted F1 values for fine-grained classifications into the four hostile classes. In all classes, (except for `Defamation' where a 0.16\% performance drop is seen for the Weighted F1 metric) the classifier performance is enhanced upon introducing the Task Adaptive Pretraining. In table \ref{taskres}, we present our official results with team name `iREL IIIT’ on the blind test corpora and compare it to the first and second runner-ups of the shared task.

\section{Experimental Details}

We used AI4Bharat’s official implementation of IndicBERT\footnote{\url{github.com/AI4Bharat/indic-bert}} as part of Hugging Face’s\footnote{\url{https://huggingface.co/}} Transformers library. All of our experimentation code was written using pytorch. We considered maximum input sequence length of 128 for both of our Transformer models, namely, the cleaned text encoder and the hashtag flow encoder. Transformer weights of both of these encoders were jointly tuned during the fine-tuning phase. We used AllenAI’s implementation\footnote{\url{github.com/allenai/dont-stop-pretraining}} of Task Adaptive Pretraining based on the Masked Language Modeling objective. The continued pretraining of IndicBERT upon the curated task specific text was performed for 100 epochs with other hyperparameters set to their default values. The cleaned text encoder was initialized with these Transformer weights before the fine-tuning phase.

For fine-tuning our end-to-end architecture, we used Adam \cite{adam} optimizer with a learning rate of 1e-5 and a dropout \cite{dropout} probability value of 0.1. All other hyperparameters were set to their default values and the fine-tuning was continued for 10 epochs. We saved model weights at the ends of each epoch and utilized the set of weights yielding the best macro F1 score on the validation set. The same schema of training and model weight saving was adopted for the coarse binary hostility detector as well as the four binary classification models for hate, defamation, offensive and fake posts.

\section{Conclusion}

In this paper, we have presented a state-of-the-art hostility detection system for Hindi Tweets. Our model architecture utilizing IndicBERT as the base Transformer, which is aware of features relevant to social media style of text in addition to the clean textual information is capable of both identifying hostility within Tweets and performing a fine-grained multi-label classification to place them into the buckets of hateful, defamation, offensive and fake. Our studies proved the efficacy of performing Task Adaptive Pretraining (TAPT) of Transformers before using such models as components of a to-be fine-tuned architecture. We experimentally showed 1.35\% and 1.40\% gains for coarse hostility detection and average gains of 4.06\% and 1.05\% for the four types of binary classifications, on macro and weighted F1 score metrics respectively in both cases. Our system ranked first in the `Hostile Post Detection in Hindi’ shared task with an F1 score of 97.16\% for coarse-grained detection and weighted F1 score of 62.96\% for fine-grained classification on the provided blind test corpora.

\bibliographystyle{splncs04}
\bibliography{constraint}

\end{document}